\documentclass{article}

\PassOptionsToPackage{sort&compress,numbers}{natbib}
\usepackage[preprint]{neurips_2025}

\usepackage[pagebackref=true,unicode=true,colorlinks=true,citecolor=blue,linkcolor=red]{hyperref}
\usepackage[utf8]{inputenc} 
\usepackage[T1]{fontenc}    
\usepackage{hyperref}       
\usepackage{url}            
\usepackage{booktabs}       
\usepackage{enumitem}      
\usepackage{enumitem}      
\usepackage{amsfonts}       
\usepackage{nicefrac}       
\usepackage{microtype}      
\usepackage{algorithm}
\usepackage{pifont}
\usepackage{float} 
\usepackage{amsmath,amsfonts}
\usepackage{algpseudocode}
\usepackage{array}
\usepackage[caption=false,font=normalsize,labelfont=sf,textfont=sf]{subfig}
\usepackage{textcomp}
\usepackage{stfloats}
\usepackage{url}
\usepackage{verbatim}
\usepackage{graphicx}
\usepackage{booktabs}
\usepackage{multirow}
\usepackage{algorithmicx}
\usepackage{adjustbox}
\usepackage{enumitem}
\usepackage[table]{xcolor}

\title{DM$^3$Net: Dual-Camera Super-Resolution via Domain Modulation and Multi-scale Matching}

\author{
\textbf{Cong Guan\textsuperscript{1,*}},
\textbf{Jiacheng Ying\textsuperscript{2,*}},
\textbf{Yuya Ieiri\textsuperscript{1}},
\textbf{Osamu Yoshie\textsuperscript{1,$\dagger$}} \\
\textsuperscript{1}Waseda University, Japan \quad
\textsuperscript{2}Zhejiang University, China \\
\textsuperscript{*}Equal contribution \quad
\textsuperscript{$\dagger$}Corresponding author \\
}

\begin{document}

\newcommand{\netname}{DM$^3$Net}
\maketitle

\begin{figure}[H]
    \centering
    \includegraphics[width=0.9\linewidth]{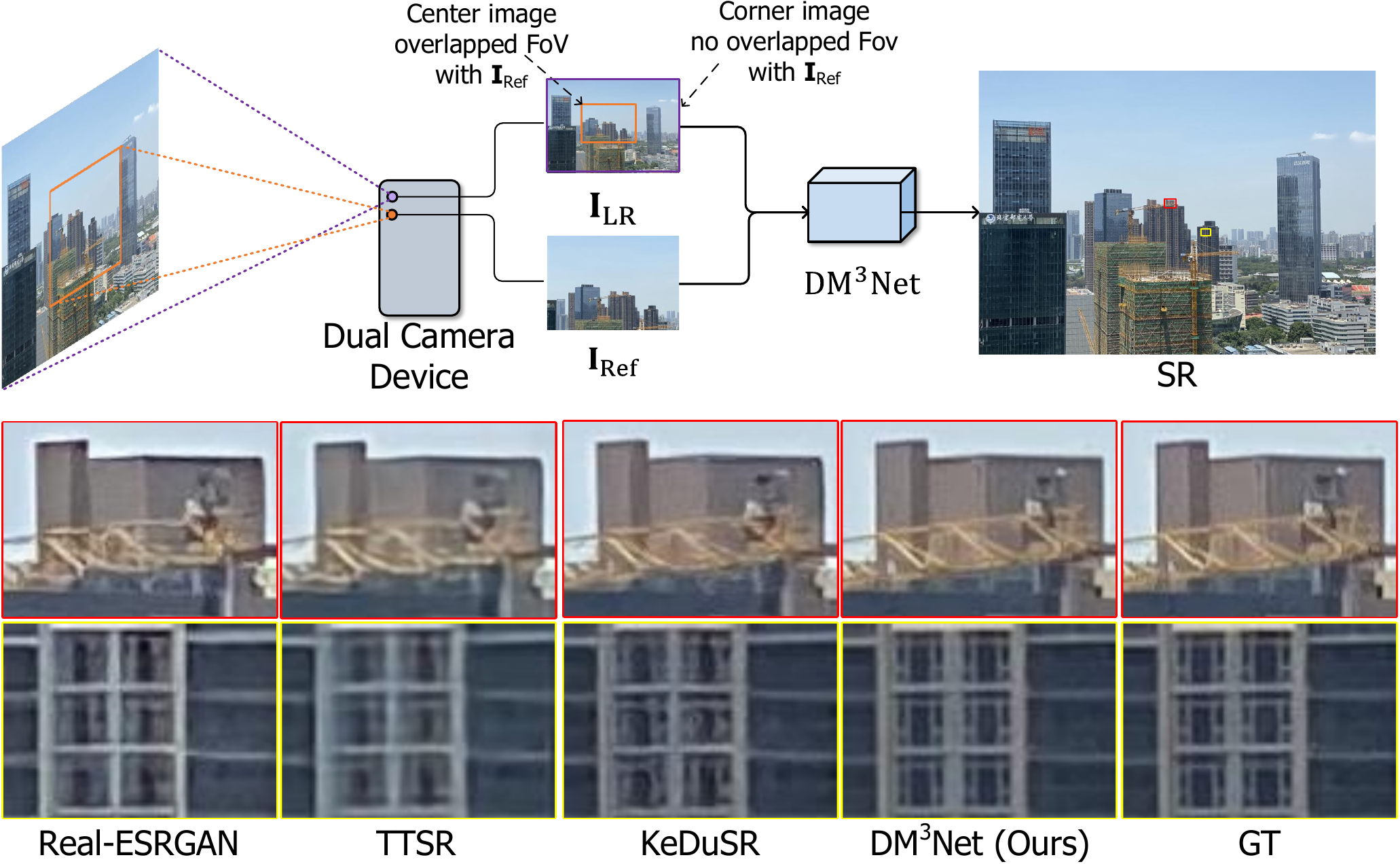}
    \caption{Modern multi-lens devices capture scenes at multiple focal lengths simultaneously. By fusing images from these diverse focal ranges, overall image quality and visual appeal can be substantially improved. The proposed method, {\netname}, achieves state-of-the-art performance with clear advantages in perceptual quality.}
    \label{fig:example_1}
\end{figure}

\begin{abstract}
  Dual-camera super-resolution is highly practical for smartphone photography that primarily super-resolve the wide-angle images using the telephoto image as a reference. In this paper, we propose DM$^3$Net, a novel dual-camera super-resolution network based on \textbf{D}omain \textbf{M}odulation and \textbf{M}ulti-scale \textbf{M}atching. To bridge the domain gap between the high-resolution domain and the degraded domain, we learn two compressed global representations from image pairs corresponding to the two domains. To enable reliable transfer of high-frequency structural details from the reference image, we design a multi-scale matching module that conducts patch-level feature matching and retrieval across multiple receptive fields to improve matching accuracy and robustness. Moreover, we also introduce Key Pruning to achieve a significant reduction in memory usage and inference time with little model performance sacrificed.  Experimental results on three real-world datasets demonstrate that our DM$^3$Net outperforms the state-of-the-art approaches.
\end{abstract}
\section{Introduction}
\vspace{-2mm}

With the advancement of multi-camera device hardware (e.g., smartphones and action cameras), it has become possible to simultaneously capture images at different focal lengths and resolutions using both wide-angle and telephoto lenses on the same device. The wide-angle lens produces images with a larger FOV (Field of View) but lower resolution, while the telephoto lens captures a narrower FOV with higher resolution. The telephoto image, which contains high-frequency details, can serve as a reference for the super-resolution of wide-angle images. This task is a sub-problem in the Reference-based Super-Resolution (RefSR) field, and is referred to as Dual-Camera Super-Resolution~\cite{wang2021dualcam}.

Due to the content inconsistency between telephoto and wide-angle images, previous works have exploited patch matching mechanisms to transfer high-fidelity patches from the reference image to the target image. For instance, DCSR~\cite{wang2021dualcam} performs feature matching between the reference and LR images to locate the most relevant reference patch for each LR patch, enabling detail fusion. Similarly, KeDuSR~\cite{yue2024kedusr} matches the LR and LR-center images to locate the optimal patch positions, and uses the corresponding reference patches to reassemble a high-resolution feature. However, the matching strategies adopted by previous works are generally implemented at a coarse scale, which limits the receptive field and leads to a loss of fine details during reconstruction.

To resolve this, we propose a multi-scale matching mechanism. Specifically, we extract multi-scale features from the LR and LR-center images as the Query ($\mathbf{Q}$) and Key ($\mathbf{K}$) features. After applying feature patchification, we perform $\mathbf{Q}$-$\mathbf{K}$ patch matching and reference feature reassembly at each scale. The resulting high-resolution features from all scales are then fused to enhance the final reconstruction. Moreover, since many key patches represent highly similar regions and cause redundant matching operations, we introduce a Key Pruning strategy to suppress unnecessary computations. This significantly reduces inference time and memory usage with minimal degradation in performance.

In addition to the local matching strategy for detail refinement, we further exploit a global domain modulation mechanism. We learn two global domain-aware embeddings from the LR-GT image pair and the LR center–Ref image pair to model the global domain gap between the high-resolution and degraded domains. Since the two embeddings are extracted from different image contents, we introduce a domain-aware loss to encourage their distribution similarity to filter out the content impact. These embeddings are then used to modulate the image reconstruction process, effectively improving the reconstruction accuracy.

Built upon the above multi-scale matching and domain modulation mechanisms, we propose a novel dual-camera super-resolution network, named DM$^3$Net. Fig.\ref{fig:example_1} presents example results generated by Real-ESRGAN~\cite{wang2021realesrgan}, TTSR~\cite{yang2020ttsr}, KeDuSR~\cite{yue2024kedusr}, and our DM$^3$Net. It can be observed that our method achieves noticeably better accuracy in reconstructing fine details compared to the others.

Our contributions can be summarized as follows:

\begin{itemize}[labelsep=0.5em, leftmargin=1.5em]
    \item We propose {\netname}, a novel dual-camera super-resolution network based on domain modulation and multi-scale matching. Extensive experiments on three datasets validate that the proposed {\netname} outperforms the existing approaches in both center region and corner region. Moreover, the proposed {\netname} also has the best generalization ability in cross-dataset evaluation.

    \item To transfer high-fidelity details from reference image, we design a multiscale matching module that search the best-matching reference patches across different receptive fields for high-resolution feature reassembly and fusion. To improve the model efficiency, we introduce Key Pruning that eliminates redundant key patches during the matching process, significantly reducing both memory usage and inference time with little performance degradation.
    
    \item To globally bridge the domain gap between the high-resolution domain and the degraded domain, we introduce a domain modulation mechanism that learns two global domain-aware embeddings from the LR-GT image pair and the LR center-Ref image pair. Each of the embeddings is used to modulate the image reconstruction process, effectively improving the reconstruction accuracy.

\end{itemize}
 
\begin{figure*}[t]
    \centering
    \includegraphics[width=1\linewidth]{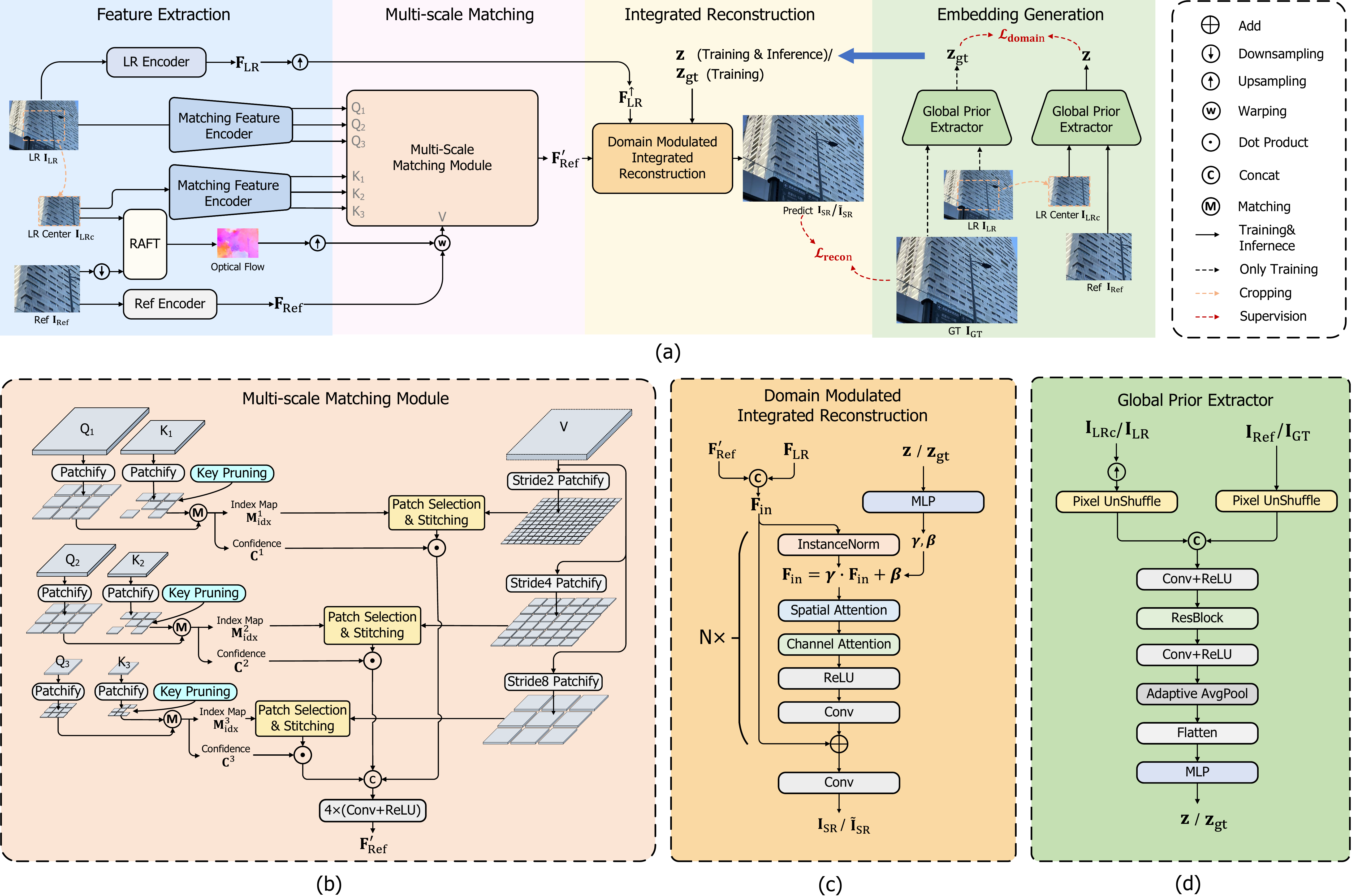}
    \vspace{-5mm}
    \caption{Schematics and detailed architectures of \netname. (a) The framework of the proposed {\netname} for dual camera super-resolution. The pipeline consists of four image encoders for feature extraction, the RAFT module for dense registration, the multi-scale matching (MSM) module for feature matching and reassembly, the global prior extractors (GPE) for domain-aware embeddings generation, and the domain modulated integrated reconstruction (DMIR) for resultant image prediction. (b) Detailed structure of multi-scale matching module. (c) Details of domain modulated integrated reconstruction. (d) Details of global prior extractor.}
    \label{fig:framework}
    \vspace{-3mm}
\end{figure*}

\vspace{-4mm}
\section{Related Work}
\vspace{-2mm}
\subsection{Reference-based Super-Resolution (RefSR)}
Reference-based SR (RefSR) methods aim to super-resolve an LR image by utilizing an additional high-resolution reference image that shares similar content with the LR input. A major challenge in RefSR is establishing accurate correspondences between the LR and the reference images. Early works~\cite{zheng2018crossnet,yang2020ttsr,zhang2019srntt,jie2022DASR,Shim_2020_CVPR,landmark} employ optical flow-based warping modules to align the reference to the LR input across multiple scales; and some works, like SADC-SR~\cite{Shim_2020_CVPR}, use deformable convolution~\cite{dai2017deformable,zhu2019deformable} to get the aligned feature. In contrast, SRNTT~\cite{zhang2019srntt} and C$^2$-Matching~\cite{jiang2021c2matching} perform patch-wise matching in feature space to transfer high-frequency details, rendering them more robust to misalignment. Transformer-based models such as TTSR~\cite{yang2020ttsr, cao2022datsr}  and MASA-SR~\cite{lu2021masasr} further improve this approach by using multi-head attention~\cite{vaswani2017attention} mechanisms to adaptively fuse relevant features. Additionally, works like MASA-SR~\cite{lu2021masasr} and AMSA~\cite{xia2022coarse} introduce strategies based on coarse-to-fine matching and contrastive learning to accelerate the matching process and enhance correspondence accuracy. These methods clearly demonstrate that effective texture transfer from a well-aligned reference image can significantly boost SR performance, assuming that a suitable high-resolution reference is available.

\vspace{-2mm}
\subsection{Dual-Camera Super-Resolution}
\vspace{-2mm}
Dual-camera super-resolution is highly practical for mobile photography, where images are simultaneously captured using different lenses. Typically, a wide-angle camera produces a low-resolution image covering a large FoV, while a telephoto camera captures a higher-resolution image for a narrower central region~\cite{dong2016image}. In this scenario, the telephoto image serves as the reference to enhance the wide-angle LR image. Traditional approaches~\cite{park2016brightness, rublee2011orb, guo2020lowlight} in dual-camera systems often involve brightness and color correction along with global registration to mitigate differences between the images.

More recently, deep learning methods have been introduced to address the unique challenges of dual-camera SR, including partial overlap, misalignment, and domain discrepancies due to different imaging pipelines. For instance, DCSR ~\cite{wang2021dualcam} is one of the first deep learning networks tailored for real-world dual-camera super-resolution. DCSR not only leverages the high-frequency details from the telephoto image but also integrates alignment modules to adjust for spatial and color differences between the two images. In addition, other approaches such as SelfDZSR~\cite{zhang2022selfdzsr} and ZeDuSR~\cite{xu2023zedusr} adopt self-supervised frameworks, wherein the telephoto image provides the supervisory signal, thus bypassing the need for externally provided HR images. Moreover, efficient systems design~\cite{wu2023hybridzoom} and specialized matching strategies~\cite{yue2024kedusr} further advance dual-camera SR by addressing efficiency and robustness under practical constraints.


\vspace{-2mm}
\section{Method}
\vspace{-2mm}
\subsection{Overview}

Fig. \ref{fig:framework} (a) illustrates the framework of the proposed {\netname}, which consists of four stages, \emph{i.e.} feature extraction, multi-scale matching, embedding generation, and integrated reconstruction. Given a wide-angle image $\mathbf{I}_{\rm LR}\in \mathbb{R}^{H\times W \times 3}$ and a telephoto reference image $\mathbf{I}_{\rm Ref}\in \mathbb{R}^{H\times W \times 3}$, we first crop the LR center region $\mathbf{I}_{\rm LRc}\in \mathbb{R}^{h\times w \times 3}$ that has the same FOV with $\mathbf{I}_{\rm Ref}$. The spatial dimensions are related with the scale factor $d$ as $H=d \times h$ and $W=d \times w$. The target of {\netname} is to predict a HR image $\mathbf{I}_{\rm SR}\in \mathbb{R}^{dH\times dW \times 3}$ with richer structure details.

To achieve this, we extract features from $\mathbf{I}_{\rm LR}$, $\mathbf{I}_{\rm LRc}$, and $\mathbf{I}_{\rm Ref}$, separately. Optical flow between $\mathbf{I}_{\rm LRc}$ and $\mathbf{I}_{\rm Ref}$ is estimated to guide the feature alignment. Multi-scale matching is then performed between $\mathbf{I}_{\rm LR}$ and $\mathbf{I}_{\rm LRc}$ based on patch similarity to search the best-matching reference patches for high-resolution feature reassembly. In parallel, two global domain-aware embeddings $\mathbf{z}$ and $\mathbf{z}_{\rm gt}$ are extracted from the $\mathbf{I}_{\rm LRc}$-$\mathbf{I}_{\rm Ref}$ pair and $\mathbf{I}_{\rm LR}$-$\mathbf{I}_{\rm GT}$ pair to modulate the reconstruction process. These two embeddings are encouraged to follow a similar distribution through a domain-aware loss during training. Finally, the LR features $\mathbf{F}_{\rm LR}^{\uparrow}$, the Ref feature $\mathbf{F}_{\rm Ref}^{'}$ and the domain-aware embeddings are fed into a reconstruction module to generate the output image $\mathbf{I}_{\rm SR}$.





\vspace{-2mm}
\subsection{Feature Extraction}

\noindent
\textbf{LR Encoder and Ref Encoder:} The LR encoder and Ref encoder share a similar architecture, each consisting of several residual blocks~\cite{he2016deep} and channel attention modules~\cite{8578843}. These encoders are designed to extract the LR feature $\mathbf{F}_{\rm LR}\in\mathbb{R}^{C_1\times H\times W}$ and the Ref feature $\mathbf{F}_{\rm Ref}\in\mathbb{R}^{C_1\times H\times W}$ from $\mathbf{I}_{\rm LR}$ and $\mathbf{I}_{\rm Ref}$, respectively. The extracted LR feature $\mathbf{F}_{\rm LR}$ is then upsampled to $\mathbf{F}_{\rm LR}^{\uparrow}\in\mathbb{R}^{C_1\times dH\times dW}$ to match the target resolution.

\noindent
\textbf{RAFT:} To address the misalignment between $\mathbf{I}_{\rm LRc}$ and $\mathbf{I}_{\rm Ref}$, we employ a pretrained RAFT model (small version)~\cite{teed2020raft} to estimate the optical flow $\mathbf{O}\in\mathbb{R}^{2\times h\times w}$ at a coarse scale. The estimated optical flow is then upsampled to $H \times W$ and used to warp the Ref feature $\mathbf{F}_{\rm Ref}$, resulting in the aligned feature $\mathbf{V}\in\mathbb{R}^{C_1\times H\times W}$: 
\begin{equation}
\mathbf{V} = \mathtt{Warp}(\mathbf{F}_{\text{Ref}}, \mathbf{O}\uparrow),
\end{equation}
where $\uparrow$ denotes the upsampling operation. We note that the RAFT module is unfrozen during training to participate in the joint optimization.

\noindent
\textbf{Matching Feature Encoder:} We use the pretrained VGG-19~\cite{simonyan2015very} network as the matching feature encoder and take the three hidden layers as the multi-scale features. We extract the multi-scale $\mathbf{Q}$ and $\mathbf{K}$ features from $\mathbf{I}_{\rm LR}$ and $\mathbf{I}_{\rm LRc}$, respectively, using the weight-shared matching feature encoders, formulated as
\begin{equation}
    \begin{aligned}
    \mathbf{Q}_{1},\mathbf{Q}_{2},\mathbf{Q}_{3} &= \mathtt{MFE}(\mathbf{I}_{\rm LR}), \\
    \mathbf{K}_{1},\mathbf{K}_{2},\mathbf{K}_{3} &= \mathtt{MFE}(\mathbf{I}_{\rm LRc}).
    \end{aligned}
\end{equation}
where $\mathbf{Q}_{1}\in\mathbb{R}^{C_2\times H\times W}$, $\mathbf{Q}_{2}\in\mathbb{R}^{C_2\times \frac{H}{2}\times \frac{W}{2}}$, $\mathbf{Q}_{3}\in\mathbb{R}^{C_2\times \frac{H}{4}\times \frac{W}{4}}$, $\mathbf{K}_{1}\in\mathbb{R}^{C_2\times h\times w}$, $\mathbf{K}_{2}\in\mathbb{R}^{C_2\times \frac{h}{2}\times \frac{w}{2}}$, $\mathbf{K}_{3}\in\mathbb{R}^{C_2\times \frac{h}{4}\times \frac{w}{4}}$, and $\mathtt{MFE(\cdot)}$ denote the matching feature encoder.



\begin{figure*}[t]
    \centering
    \includegraphics[width=0.95\linewidth]{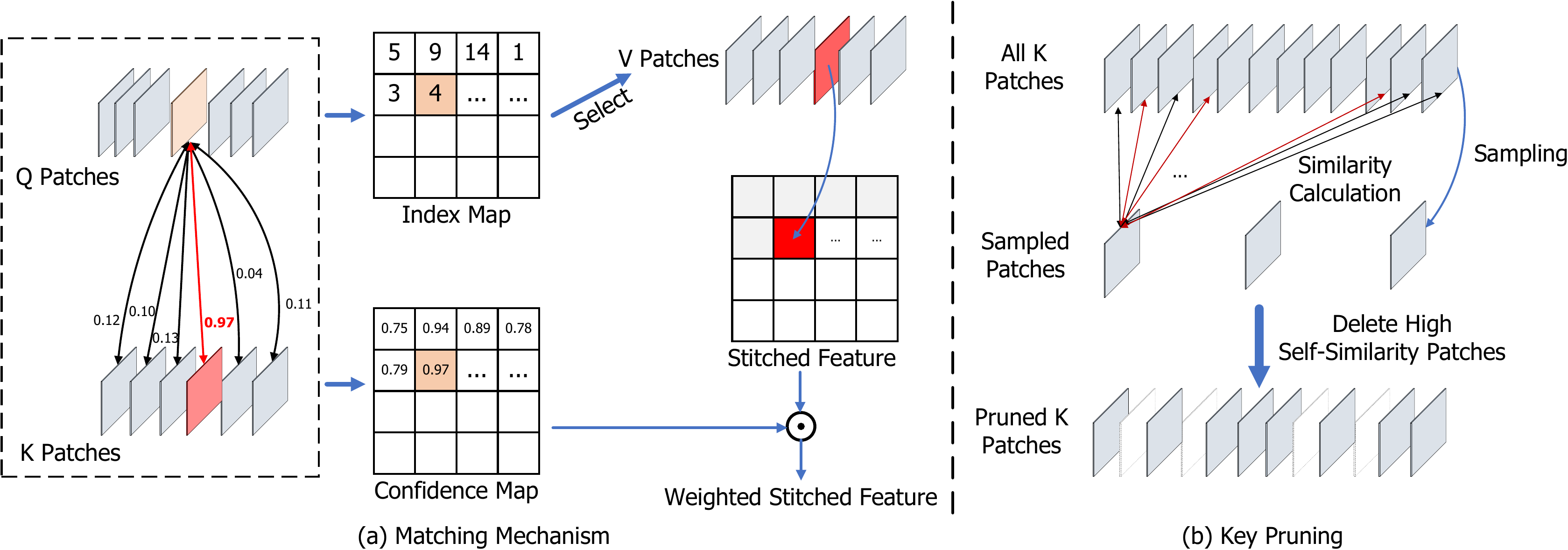}
    \caption{The flow of (a) the Matching Mechanism and (b) the Key Pruning.}

    \label{fig:matching}
\end{figure*}

\vspace{-2mm}
\subsection{Multi-scale Matching}

\begin{figure*}[t]
    \centering
    
    \includegraphics[width=0.95\linewidth]{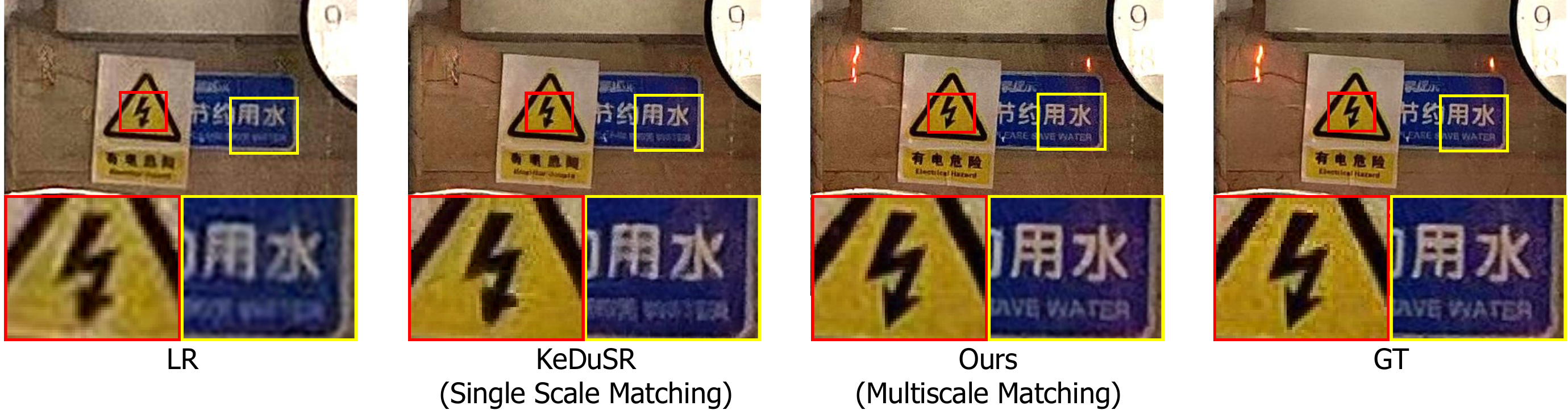}
    \vspace{-3mm}
    \caption{Reconstructed images produced by the single-scale matching approach, KeDuSR\cite{yue2024kedusr}, and our proposed method. Our method achieves better preservation of texture details.}
    \label{fig:scale-compare}
    \vspace{-3mm}
\end{figure*}


The feature of LR center region can be complemented by the aligned high-resolution Ref features. However, in the corner regions, it is difficult to directly obtain corresponding high-resolution features. To address this issue, previous approaches~\cite{yue2024kedusr,wang2021dualcam} adopt block matching strategy to search similar feature patches in the center region and use them to enhance the feature patches in the corner region. However, these methods typically perform single-scale feature matching at a coarse scale, which restricts the receptive field of the matching process. In this work, we propose a multi-scale matching module to further improve the accuracy and robustness of feature transfer from the Ref image. As illustrated in Fig.~\ref{fig:framework} (b), the multi-scale matching module takes the $\mathbf{V}$ feature, multi-scale $\mathbf{Q}$ and $\mathbf{K}$ features as the input. For the $i$-th scale, we patchify $\mathbf{V}$, $\mathbf{Q}_{i}$ and $\mathbf{K}_{i}$ to obtain patches of V, Q and K. And then, we compute cosine similarity between each Q patches and all K patches to select the best‐matching K patch to form both a matching index map $\mathbf{M}_{\text{idx-}i}$ and a confidence map $\mathbf{C}_{i}$, as illustrated in Fig.~\ref{fig:matching} (a). Each element of $\mathbf{M}_{\text{idx-}i}$ and $\mathbf{C}_{i}$ can be computed as
\begin{equation}
    {M}_{\text{idx-}i}^{j} = \arg\max_{1\le k\le N}\;\mathtt{sim}(\mathbf{Q}_{i}^{j}, \mathbf{K}_i^k),
\end{equation}
\begin{equation}
    {C}_{i}^{j} = \max_{1\le k\le N}\;\mathtt{sim}(\mathbf{Q}_i^{j}, \mathbf{K}_i^k)
\end{equation}
where $j$ denotes the element position, $N$ denotes the number of K patches, $\mathtt{sim}(\cdot)$ denotes the cosine similarity operator. The index map $\mathbf{M}_{\text{idx-}i}$ is then used to retrieve the corresponding V patches and reassemble a high-resolution feature, while the confidence map $\mathbf{C}_{i}$ is applied via element-wise multiplication to adaptively weight the stitched features. Finally, we concatenate the three weighted feature maps and feed them to a fusion head that consists of four convolution and ReLU\cite{agarap2018deep} layers to produce the output feature $\mathbf{F}_{\mathrm{Ref}}^{'}\in \mathbb{R}^{C_1\times dH\times dW}$. 

Fig~\ref{fig:scale-compare} presents an example image generated by KeDuSR~\cite{yue2024kedusr}, which employs single-scale matching, and our method. It is observed our method 
outperforms KeDuSR in terms of structural consistency and fine details. This improvement is primarily attributed to the enhanced reliability of feature transfer enabled by our multi-scale matching mechanism.

\begin{table}[t]
    \centering
    \caption{The performance, inference time, and memory usage before and after using Key Pruning. Tests are conducted on NVIDIA H20 on CameraFusion-Real dataset.}
    \label{tab:Keypruning}
    \begin{adjustbox}{width=0.8\linewidth} 
    \begin{tabular}{l|ccc}
    \toprule
    {} &\textbf{PSNR$\uparrow$ / SSIM $\uparrow$}   & \textbf{Inference Time (s)}    & \textbf{Memory Usage (GB)}    \\
    \midrule
    Before Key Pruning & 27.87 / 0.8436 & 31.08 & 49.41 \\
    
    After Key Pruning  &    27.75 / 0.8364     & 11.32   &  40.95  \\

    \bottomrule
    \end{tabular}
    \end{adjustbox}
    \vspace{-5mm}
\end{table}

However, we notice that the above multi-scale matching mechanism between the $\mathbf{Q}$ and $\mathbf{K}$ features incurs significant memory consumption and increases inference time. Considering that many K patches are highly similar to some others, a large portion of the similarity computations between Q and K patches are redundant and unnecessary. Motivated by this observation, we propose a simple yet effective strategy, Key Pruning, to improve the efficiency of the model, as illustrated in Fig. \ref{fig:matching} (b). We uniformly sample a subset of K patches, compute the similarity between the sampled patches and all K patches, and remove those patches whose similarity exceeds a certain threshold. The remaining patches serve as the pruned K patches for subsequent matching. In this work, we set the sampling interval as 16 in both the horizontal and vertical directions, and set the threshold as 0.7. Through Key Pruning, we achieve a substantial reduction in resource consumption by sacrificing little model performance. Specifically, as listed in table~\ref{tab:Keypruning}, on the CameraFusion-Real dataset, memory usage was reduced from 49.41 GB to 40.95 GB, and inference time decreased from 31.08s to 11.32s, with only a slight performance drop of 0.12 dB in PSNR and 0.072 in SSIM.

\subsection{Embedding Generation}
\begin{figure}[h]
    \centering
    \includegraphics[width=0.5\linewidth]{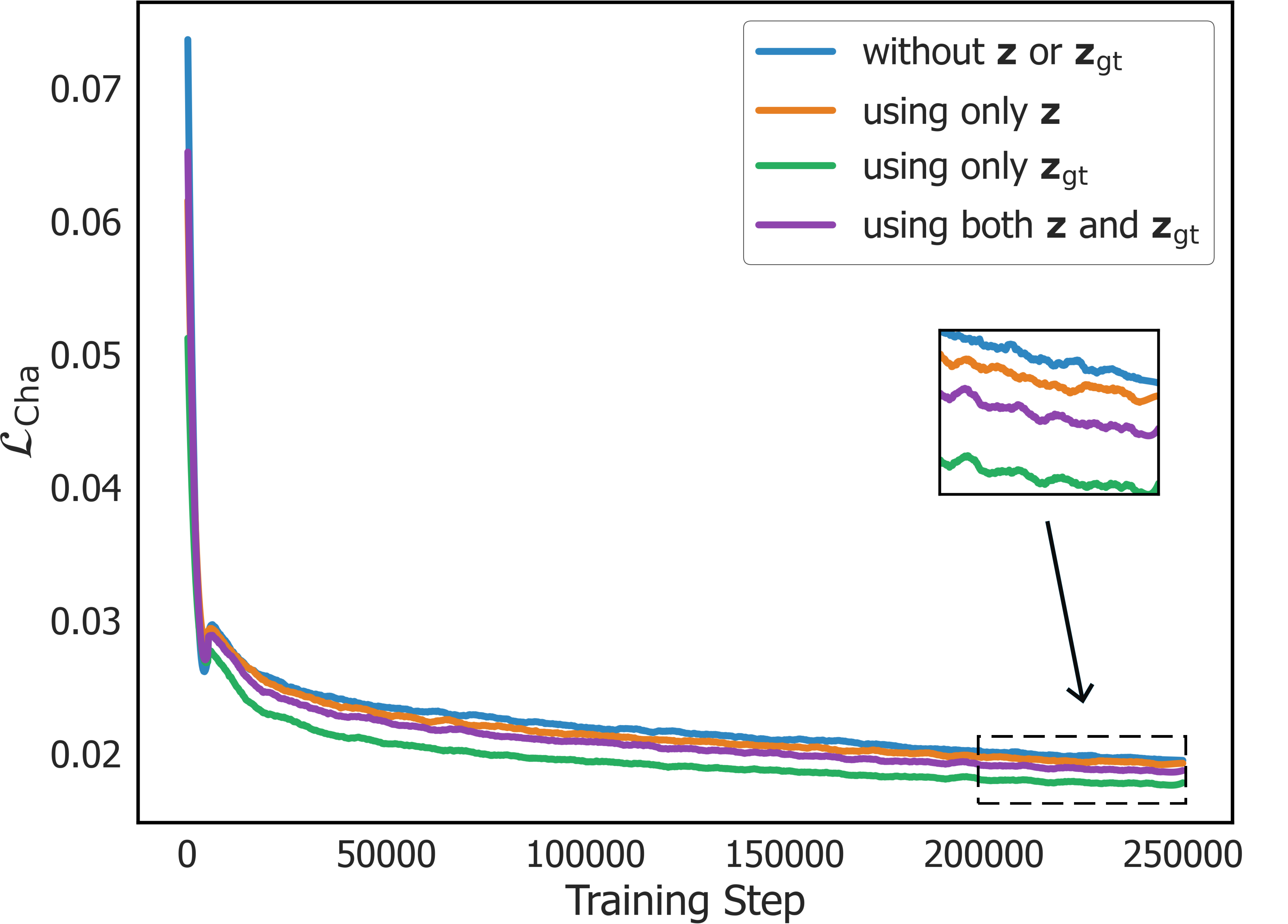}
    \vspace{-1mm}
   \caption{Visualization of the Charbonnier loss~\cite{charbonnier1994two} curves during training under four experiment settings— (1) without $\mathbf{z}$ or $\mathbf{z}_{\rm gt}$, (2) using only $\mathbf{z}$, (3) using only $\mathbf{z}_{\rm gt}$, and (4) using both $\mathbf{z}$ and $\mathbf{z}_{\rm gt}$.}
    \label{fig:losscurve}
    \vspace{-2mm}
\end{figure}

Motivated by previous image restoration approaches~\cite{xia2023diffir, thehdr}, we also learn a compressed global representation that captures the high-fidelity prior to bridge the domain gap between the high-resolution domain and the degraded domain. To this end, we propose \emph{Global Prior Extractor} (GPE) that extracts a domain‑aware embedding from two images corresponding to the two domains. The structure of GPE is illustrated in Fig.~\ref{fig:framework} (d). In particular, we consider two different input pairs: $\{\mathbf{I}_{\text{LR}}, \mathbf{I}_{\text{GT}}\}$ and $\{\mathbf{I}_{\text{LRc}}, \mathbf{I}_{\text{Ref}}\}$. Both pairs share two key characteristics: (1) the two images have the same scene region, and (2) one in the high-resolution domain and the other in the degraded domain. We employ two GPEs to extract domain-aware embeddings $z\in \mathbb{R}^{1\times1024}$ and $\mathbf{z}_{\rm gt}\in \mathbb{R}^{1\times1024}$ using the pair $\{\mathbf{I}_{\text{LRc}}, \mathbf{I}_{\text{Ref}}\}$ and pair $\{\mathbf{I}_{\text{LR}}, \mathbf{I}_{\text{GT}}\}$, respectively, formulated as 
\begin{equation}
    \mathbf{z}= \mathtt{GPE}_1 \left(\mathbf{I}_{\text{LRc}}, \mathbf{I}_{\text{Ref}} \right), 
\end{equation}
\begin{equation}
    \mathbf{z}_{\rm gt}= \mathtt{GPE}_2 \left(\mathbf{I}_{\text{LR}}, \mathbf{I}_{\text{GT}} \right). 
\end{equation}
Each of the embeddings is employed in the Integrated Reconstruction stage for image reconstruction. 

Since $\mathbf{z}_{\rm gt}$ is extracted from the GT image, the Charbonnier loss, one of the image reconstruction loss, of the model using $\mathbf{z}_{\rm gt}$ during training is significantly lower than that using $\mathbf{z}$ (see the orange and blue curves in Fig.~\ref{fig:losscurve}). However, $\mathbf{z}_{\rm gt}$ is unavailable during inference because the GT image is not accessible. To mitigate this gap, we employ the domain-aware loss $\mathcal{L}_{\rm domain}$ to enforce $\mathbf{z}$ to have similar distribution to $\mathbf{z}_{\rm gt}$, which also helps reduce the image reconstruction loss (see the green curve in Fig.~\ref{fig:losscurve}). In this way, during inference, we can use $\mathbf{z}$ as a substitute for $\mathbf{z}_{\rm gt}$ for image reconstruction.

\vspace{-2mm}
\subsection{Integrated Reconstruction}

In this stage, the reassembled ref feature $\mathbf{F}_{\rm Ref}^{'}$, the LR feature $\mathbf{F}_{\rm LR}^{\uparrow}$, and the domain-aware embedding $\mathbf{z} / \mathbf{z_{\rm gt}}$ are fed into the Domain Modulated Integrated Reconstruction (DMIR) module to predict the super-resolved image $\mathbf{I}_{\rm SR}/\tilde{\mathbf{I}}_{\rm SR}$. The architecture of DMIR module is illustrated in Fig.~\ref{fig:framework} (c), where the domain-aware embedding $\mathbf{z}/\mathbf{z}_{\rm gt}$ is processed by an MLP\cite{haykin1994neural} to generate affine parameters $\gamma$ and $\beta$ for feature modulation~\cite{Huang_2017_ICCV}.


We note that during the training phase, $\mathbf{I}_{\rm SR}$ denotes the output using $\mathbf{z}$ as the input, while $\tilde{\mathbf{I}}_{\rm SR}$ denotes the output using $\mathbf{z}_{\rm gt}$. Both outputs are supervised to encourage $\mathbf{z}$ and $\mathbf{z_{\rm gt}}$ to effectively bridge the high-resolution and degraded domain gap. During inference, only $\mathbf{z}$ is required as input to generate the predicted image $\mathbf{I}_{\rm SR}$.



\vspace{-2mm}
\subsection{Loss Functions}



%
In the training phase, we employ reconstruction loss $\mathcal{L}_{\text{rec}}$ and domain-aware loss $\mathcal{L}_{\text{domain}}$ to guide the learning process. The reconstruction loss consists of the Charbonnier loss $\mathcal{L}_{\text{cha}}$ and the perceptual loss $\mathcal{L}_{\text{per}}$, while the domain-aware loss $\mathcal{L}_{\text{domain}}$ is used to encourage the two domain-aware embeddings to have similar distributions. Please refer to Appendix for more details.


\vspace{-2mm}
\section{Experiments}
\vspace{-2mm}
\subsection{Experimental Settings}
 

\noindent
\noindent
\textbf{Datasets.} We conduct our experiments on 3 publicly available real-world dual-camera super-resolution datasets: \textbf{DuSR-Real}, \textbf{RealMCVSR-Real}, and \textbf{CameraFusion-Real}. These datasets differ in terms of degradation levels, alignment quality, and resolution. DuSR-Real provides well-aligned image triplets captured simultaneously with iPhone 13 dual-lens cameras, covering diverse scenes at a resolution of $1792\times896$. RealMCVSR-Real features more challenging degradations such as motion blur, while CameraFusion-Real offers the highest resolution ($3584\times2560$) but includes minor misalignment. We adopt these datasets to evaluate the generalization and robustness of our model under various real-world scenarios.

\noindent
\textbf{Implementation Details.} During training stage, we apply random flipping and $90^\circ$ rotations for data augmentation. The batch size is fixed to 4, and the LR patch size is set to $128 \times 128$. The model is trained for 400 epochs using the Adam optimizer~\cite{kingma2014adam} with $\beta_1 = 0.9$ and $\beta_2 = 0.999$. The initial learning rate is $1 \times 10^{-4}$ and is decayed to $5 \times 10^{-5}$ after 250{,}000 iterations. All experiments are implemented using PyTorch~\cite{paszke2019pytorch} and conducted on a single NVIDIA A100 40G PCIE GPU.

\noindent
\textbf{Evaluation Metrics.} We use PSNR~\cite{hore2010psnr}, SSIM~\cite{wang2004image}, and LPIPS~\cite{zhang2018perceptual} as evaluation metrics. 

\noindent
\textbf{Comparing Methods.} We compare our method with single image SR approaches, reference-based SR approaches, and dual-camera SR approaches. The single image SR approaches includes SwinIR~\cite{liang2021swinir} and Real-ESRGAN~\cite{wang2021realesrgan}. The reference-based SR includes MASA-SR~\cite{lu2021masasr} and TTSR~\cite{yang2020ttsr}. The dual-camera SR approaches includes DCSR~\cite{wang2021dualcam}, SelfDZSR~\cite{zhang2022selfdzsr}, and KeDuSR~\cite{yue2024kedusr}. For a fair comparison, we retrain all the aforementioned approaches using the same training set as our method.

\begin{table}[t]
  \footnotesize
  \centering
  \caption{Quantitative comparisons on DuSR-Real dataset. The best metrics are in bold. }
  \label{tab:dusr_comparison1}
  \begin{adjustbox}{width=0.8\linewidth}   
  \begin{tabular}{lcccccc}
    \toprule
    \multirow{2}{*}{Method}  & \multicolumn{3}{c}{Full-Image} & Center-Image & Corner-Image & Inference\\
    \cmidrule(lr){2-4} \cmidrule(lr){5-5} \cmidrule(lr){6-6}
    & PSNR $\uparrow$ & SSIM $\uparrow$ & LPIPS $\downarrow$ & PSNR/SSIM & PSNR/SSIM & Time (s)\\
    \midrule
    SwinIR~\cite{liang2021swinir}  &  26.02 & 0.8580 & 0.226 & 26.22 / 0.8592 & 25.99 / 0.8576 & 4.609\\
    Real-ESRGAN~\cite{wang2021realesrgan}  & 24.19 & 0.8339  & 0.188   &  24.39 / 0.8372    &      24.17 / 0.8328  &0.113   \\
    \midrule
    MASA-SR~\cite{lu2021masasr}  & 26.17 & 0.8528 & 0.233 &   26.69 / 0.8554 & 26.05 / 0.8519 &7.067\\
    TTSR~\cite{yang2020ttsr}   & 25.59 & 0.8544 & 0.234 &   26.14 / 0.8612 & 25.46 / 0.8521 & 6.013\\
    \midrule
    
    DCSR~\cite{wang2021dualcam} & 26.06  & 0.8576 &  0.190 & 28.33 / 0.8975  &   25.54 / 0.8440 &1.172 \\
    SelfDZSR~\cite{zhang2022selfdzsr} & 25.62 & 0.8351 & 0.198 &  26.23 / 0.8387  & 25.47 / 0.8339 & 0.793 \\
    KeDuSR~\cite{yue2024kedusr} & 27.18 & 0.8752 & 0.164 & 29.06 / 0.9219 & 26.77 / 0.8593 & 0.836 \\
    \midrule
    \textbf{\netname}   & \textbf{27.74} & \textbf{0.8879} & \textbf{0.159} & \textbf{29.67} / \textbf{0.9291} & \textbf{27.34} / \textbf{0.8738} & 1.404 \\
    \bottomrule
  \end{tabular}
  \end{adjustbox}
  \vspace{-3mm}
\end{table}

\begin{table}[t]
  \footnotesize
  \centering
  \caption{Quantitative comparisons on RealMCVSR-Real dataset. The best metrics are in bold.}
  \label{tab:realmcvsr_comparison1}
\begin{adjustbox}{width=0.8\linewidth} 
  \begin{tabular}{lcccccc}
    \toprule
    \multirow{2}{*}{Method} & \multicolumn{3}{c}{Full-Image} & Center-Image & Corner-Image \\
    \cmidrule(lr){2-4} \cmidrule(lr){5-5} \cmidrule(lr){6-6}
    & PSNR $\uparrow$ & SSIM $\uparrow$ & LPIPS $\downarrow$ & PSNR/SSIM & PSNR/SSIM \\
    \midrule
    SwinIR~\cite{liang2021swinir}  & 25.78 & 0.7992 & 0.304 &   25.57 / 0.7896 & 25.92 / 0.8025   \\
    Real-ESRGAN~\cite{wang2021realesrgan} & 24.00 & 0.7704  & 0.251 &  23.79 / 0.7640  & 24.15 / 0.7725    \\
    \midrule
    MASA-SR~\cite{lu2021masasr} &25.81 & 0.7911 & 0.315 &   25.73 / 0.7821 & 25.90 / 0.7941 \\
    TTSR~\cite{yang2020ttsr}  &24.83 & 0.7889 & 0.311 &   25.63 / 0.7990 & 24.63 / 0.7855 \\
    \midrule
    DCSR~\cite{wang2021dualcam} & 25.79   & 0.8024   & 0.275 &  27.14 / 0.8524 &  25.47 / 0.7853 \\
    SelfDZSR~\cite{zhang2022selfdzsr} & 25.15 & 0.7803 & 0.276 & 25.32 / 0.7748 & 25.17 / 0.7821  \\
    KeDuSR~\cite{yue2024kedusr}& 26.42 & 0.8184 & 0.226 & 28.51 / 0.9090 & 25.95 / 0.7875 \\    

    \midrule
    \textbf{\netname}        &  \textbf{27.07} &  \textbf{0.8394} & \textbf{0.222} &  \textbf{29.27} /  \textbf{0.9197} &  \textbf{26.58} /   \textbf{0.8120} \\
    \bottomrule
  \end{tabular}
  \end{adjustbox}
  \vspace{-5mm}
\end{table}

 \begin{table}[t]
  \footnotesize
  \centering
  \caption{Quantitative comparisons on CameraFusion-Real dataset. The best metrics are in bold.}
  \label{tab:camerafusion_comparison1}
  \begin{adjustbox}{width=0.8\linewidth} 
  \begin{tabular}{lcccccc}
    \toprule
    \multirow{2}{*}{Method} & \multicolumn{3}{c}{Full-Image} & Center-Image & Corner-Image \\
    \cmidrule(lr){2-4} \cmidrule(lr){5-5} \cmidrule(lr){6-6}
    & PSNR $\uparrow$ & SSIM $\uparrow$ & LPIPS $\downarrow$ & PSNR/SSIM & PSNR/SSIM \\
    \midrule
    SwinIR~\cite{liang2021swinir} & 25.32 & 0.7926 & 0.347 & 25.90 \ 0.8000 & 25.19 \ 0.7901 \\
    Real-ESRGAN~\cite{wang2021realesrgan} & 24.37 & 0.7643  & 0.287 &  24.79 / 0.7690  &  24.30 / 0.7626   \\
    \midrule
 
    MASA-SR~\cite{lu2021masasr} & 25.92 & 0.7886 &  0.343 & 26.75 / 0.8004  &  25.74 / 0.7846    \\
    TTSR~\cite{yang2020ttsr} & 25.51 & 0.7945 & 0.376 &   25.42 / 0.7890 & 25.61 / 0.7964 \\

    \midrule
    
    DCSR~\cite{wang2021dualcam} &  24.96 & 0.7569  & 0.316  & 26.15 / 0.7898  & 24.67 / 0.7458 \\
    SelfDZSR~\cite{zhang2022selfdzsr} &25.78&0.7833&  0.280     & 26.76 / 0.7950     &  25.55 /    0.7794 \\
    KeDuSR~\cite{yue2024kedusr}& 27.00 & 0.7931 & 0.274 & 29.77 / 0.8418 & 26.43 / 0.7768 \\

    \midrule
    \textbf{\netname}  & \textbf{27.75} & \textbf{0.8364} & \textbf{0.131} & \textbf{32.22}  / \textbf{0.8894} & \textbf{27.08} / \textbf{0.8186}\\
    \bottomrule
  \end{tabular}
  \end{adjustbox}
\end{table}

\vspace{-2mm}
\subsection{Quantitative Comparison}
\vspace{-1mm}
Table~\ref{tab:dusr_comparison1}, Table~\ref{tab:realmcvsr_comparison1}, and Table~\ref{tab:camerafusion_comparison1} present the quantitative results on three benchmark datasets. Full-Image denotes evaluation over the entire image, Center-Image corresponds to the overlapping FOV region, and Corner-Image refers to the non-overlapping regions. It can be observed that, at the Full-Image level across all datasets, our {\netname} consistently achieves the best performance in all metrics, including PSNR, SSIM, and LPIPS, demonstrating that our proposed {\netname} generates the most faithful reconstruction results. Moreover, the performance advantage of our method remains evident in both Center-Image and Corner-Image regions, especially on the CameraFusion-Real dataset, highlighting its strong restoration capability and robustness to FoV variation.



 
\begin{figure*}[t]
    \centering
    \includegraphics[width=1\linewidth]{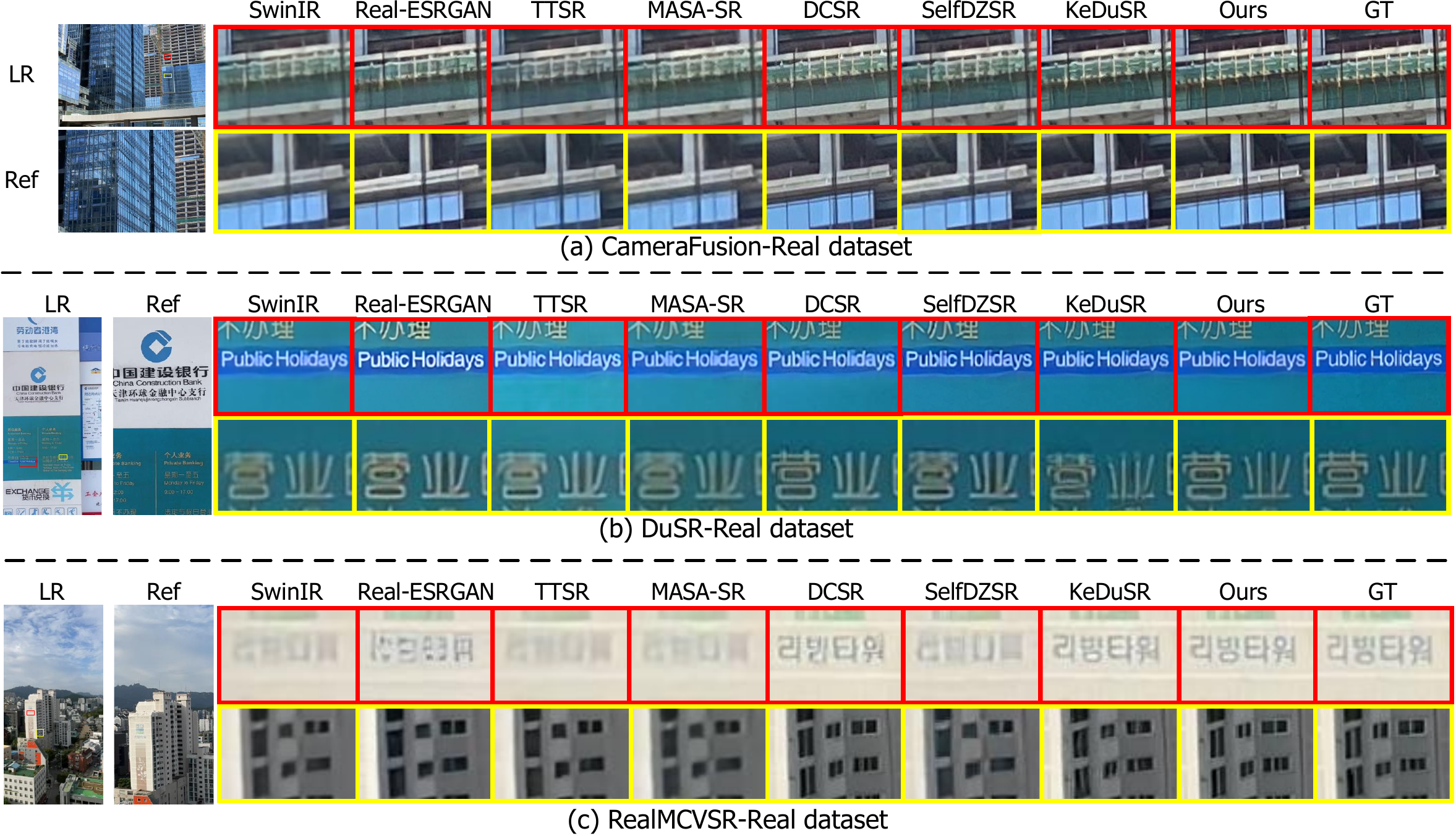}
    \vspace{-6mm}
    \caption{Qualitative comparison results on 3 datasets. Zoom in for best comparison.}
    \label{fig:qualitative-compare1}
    \vspace{-4mm}
\end{figure*}

\vspace{-2mm}
\subsection{Qualitative Comparison}
\vspace{-1mm}

Figure~\ref{fig:qualitative-compare1} presents the qualitative comparisons on three benchmark datasets. SwinIR, TTSR, MASA-SR, and SelfZeDuSR tend to produce blurred textures and distorted structures (see red boxes in CameraFusion-Real, yellow boxes in DUSR-Real, and red boxes in RealMCVSR-Real). TTSR also suffers from noticeable color distortions (e.g., samples in DuSR-Real). While Real-ESRGAN, DCSR, and KeDuSR generate sharper details, they introduce artifacts around characters (see red and yellow boxes in DuSR-Real, and red boxes in RealMCVSR-Real). KeDuSR additionally fails to preserve the correct shapes of windows and blocks (see yellow boxes in CameraFusion-Real and RealMCVSR-Real). In contrast, our  {\netname} successfully reconstructs high-frequency textures and character details without color distortion. Additional comparisons and full-resolution results are included in the supplementary material.


\definecolor{lightgray}{gray}{0.95} 

\begin{table}[t]
  \footnotesize
  \centering
  \begin{minipage}[t]{0.48\linewidth}
    \centering
    \caption{Ablation Study on Multi-scale Matching. $1/4$, $1/2$, and $1$ denote matching at respective scales.}
    \label{tab:ablation-mm}
    \begin{tabular}{ccc|cc}
      \toprule
      1/4 & 1/2 & 1 & PSNR $\uparrow$ & SSIM $\uparrow$ \\
      \midrule
      \checkmark &     &     & 26.55 & 0.8112 \\
      & \checkmark &     & 27.17 & 0.8250 \\
       &  & \checkmark    & 27.30 & 0.8327 \\
      \checkmark& \checkmark &  & 27.41 & 0.8331 \\
      \checkmark & & \checkmark & 27.55 & 0.8335 \\
      & \checkmark & \checkmark & 27.54 & 0.8365 \\
      \rowcolor{lightgray}
      \checkmark & \checkmark & \checkmark & \textbf{27.75} & \textbf{0.8364} \\
      \bottomrule
    \end{tabular}
  \end{minipage}
  \hfill
  \begin{minipage}[t]{0.48\linewidth}
    \centering
    \caption{Ablation Study on Domain Modulation.}
    \label{tab:ablation-dm}
    \begin{tabular}{cc|cc}
      \toprule
      $\mathbf{z}$ & $\mathbf{z}_{\text{gt}}$ & PSNR $\uparrow$ & SSIM $\uparrow$ \\
      \midrule
      \ding{55} & \ding{55} & 27.37 & 0.8337 \\
      \checkmark & \ding{55} & 27.50 & 0.8360 \\
      \rowcolor{lightgray}
      \checkmark & \checkmark & \textbf{27.75} & \textbf{0.8364} \\
      \bottomrule
    \end{tabular}
  \end{minipage}
  \vspace{-2mm}
\end{table}

\vspace{-2mm}
\subsection{Abalation Study}
\vspace{-1mm}
Ablation studies are conducted on the CameraFusion-Real dataset to assess the impact of the proposed multi-scale matching and domain modulation.

\textbf{Multi-Scale Matching:} To assess the impact of the multi-scale matching, we evaluated seven different combinations of matching at three scales: 1/4 scale, 1/2 scale, and full scale. As listed in Table~\ref{tab:ablation-mm}, adopting higher-scale matching leads to better reconstruction accuracy. Moreover, combining multiple scales brings further improvements, because the receptive fields associated with different scales are complementary and can mitigate the limitations of each individual scale.

\textbf{Domain Modulation:} To validate the effectiveness of the domain-aware embeddings $\mathbf{z}$ and $\mathbf{z}_{\rm gt}$, we trained three versions of models: (1) a baseline model without $\mathbf{z}$ and $\mathbf{z}_{\rm gt}$, (2) a model incorporating only $\mathbf{z}$ without supervision from $\mathbf{z}_{\rm gt}$, and (3) the full model. As listed in Table~\ref{tab:ablation-dm}, incorporating $\mathbf{z}$ alone already improves the reconstruction metrics, demonstrating that a global domain-aware embedding benefits high-resolution image reconstruction. Furthermore, the full model achieves the best performance, indicating that supervision from $\mathbf{z}_{\rm gt}$ further enhances reconstruction quality.

\vspace{-1mm}
\subsection{Generalization Evaluation}
\vspace{-1mm}
We evaluate the generalization ability of different approaches on the RealMCVSR-Real and CameraFusion-Real datasets using models trained on DUSR-Real dataset. As shown in Table~\ref{tab:generalization}, our method outperforms the competing approaches, demonstrating the superior generalization capability of {\netname}. This advantage can be attributed to our matching mechanism and the extraction of domain-aware embeddings, both of which are independent of the training dataset.

\begin{table}[t]
    \centering
    \caption{Generalization Evaluation with the models trained on DuSR-Real dataset.}
    \label{tab:generalization}
    \begin{adjustbox}{width=0.55\linewidth} 
    \begin{tabular}{lcc}
    \toprule
    \textbf{Method}
      & \textbf{RealMCVSR-Real}    & \textbf{CameraFusion-Real}    \\
    & PSNR↑ / SSIM↑             & PSNR↑ / SSIM↑                  \\
    \midrule
    SwinIR      & 24.00 / 0.7738 & 25.01 / 0.7755 \\
    Real-ESRGAN & 23.76 / 0.7680 & 23.94 / 0.7114 \\
    MASA-SR     & 25.18 / 0.7757 & 25.38 / 0.7724 \\
    TTSR        & 24.86 / 0.7796 & 24.50 / 0.7653 \\
    DCSR        & 24.96 / 0.7807 & 24.82 / 0.7530 \\
    SelfDZSR    & 24.73 / 0.7741 & 24.79 / 0.7253 \\
    KeDuSR      & 26.21 / 0.8189 & 26.78 / 0.7909 \\
    \midrule
    \textbf{\netname} 
               & \textbf{26.66 / 0.8284} & \textbf{27.26 / 0.8172} \\
    \bottomrule
    \end{tabular}
    \end{adjustbox}
    \vspace{-4mm}
\end{table}

\vspace{-2mm}
\section{Conclusions}
\vspace{-2mm}
In this work, we propose \netname, a novel framework for dual-camera super-resolution. {\netname} utilizes multi-scale matching to transfer high-quality structure details from the reference image, and leverages domain modulation to bridge the high-resolution and the degraded domain gap. By introducing a key pruning strategy, {\netname} also enhances computational efficiency with slight performance drop. Extensive experiments on three real-world datasets demonstrate that our method achieves state-of-the-art performance across multiple visual quality metrics. By explicitly modeling domain and FOV discrepancies, {\netname} achieves more accurate detail reconstruction and delivers strong generalization and robustness across diverse real-world dual-camera super-resolution scenarios.

\clearpage
\bibliographystyle{unsrt} 
\bibliography{REF}
\clearpage
\appendix

\section*{\LARGE \textbf{Appendix and supplemental material}}

\section{Loss Functions}
In the training phase, we employ reconstruction loss $\mathcal{L}_{\text{rec}}$ and domain-aware loss $\mathcal{L}_{\text{domain}}$ to guide the learning process. The reconstruction loss consists of the Charbonnier loss $\mathcal{L}_{\text{cha}}$ and the perceptual loss $\mathcal{L}_{\text{per}}$, while the domain-aware loss $\mathcal{L}_{\text{domain}}$ is used to encourage the two domain-aware embeddings to have similar distributions.

\textbf{Charbonnier Loss.}  
We use the Charbonnier loss~\cite{charbonnier1994two}, a differentiable variant of the $L_1$ loss, as the primary reconstruction loss for both predicted images ${\mathbf{I}}_{\text{SR}}$ and $\tilde{\mathbf{I}}_{\text{SR}}$. The loss is defined as:
\begin{equation}
\mathcal{L}_{\text{cha}} = \sqrt{\left(\mathbf{I}_{\text{SR}} - \mathbf{I}_{\text{GT}}\right)^2 + \epsilon^2} + \sqrt{\left(\tilde{\mathbf{I}}_{\text{SR}} - \mathbf{I}_{\text{GT}}\right)^2 + \epsilon^2},
\end{equation}
where $\epsilon$ is a small constant (set to $10^{-6}$ in our experiments) added for numerical stability.

\vspace{1mm}
\textbf{Perceptual Loss.}  
To further enhance the image visual fidelity, we introduce the perceptual loss~\cite{johnson2016perceptual} that compares deep features extracted from a pretrained VGG network~\cite{simonyan2015very}. The perceptual loss is defined as:
\begin{equation}
\mathcal{L}_{\text{percep}} = \sum_l \left\| \phi_l({\mathbf{I}}_{\text{SR}}) - \phi_l(\mathbf{I}_{\text{GT}}) \right\|_2^2 + \sum_l \left\| \phi_l(\tilde{\mathbf{I}}_{\text{SR}}) - \phi_l(\mathbf{I}_{\text{GT}}) \right\|_2^2,
\end{equation}
where $\phi_l(\cdot)$ denotes the activation of the $l$-th layer of the VGG network.

\textbf{Domain-Aware Loss.} We employ the domain-aware loss to encourage the domain-aware embeddings $\mathbf{z}$ and $\mathbf{z}_{\rm gt}$ to have similar distributions, computed as
\begin{equation}
\mathcal{L}_{\text{domain}} = \left\| \mathbf{z} - \mathbf{z}_{\text{gt}} \right\|_1.
\end{equation}

\textbf{Total Loss.}
The final training objective is formulated as a weighted combination of all loss components:
\begin{equation}
\mathcal{L}_{\text{total}} = \mathcal{L}_{\rm rec} + \lambda_{3} \mathcal{L}_{\text{domain}} = \lambda_{1} \mathcal{L}_{\text{rec}} + \lambda_{2} \mathcal{L}_{\text{per}} + \lambda_{3} \mathcal{L}_{\text{domain}},
\end{equation}
where $\lambda_{\text{rec}}$, $\lambda_{\text{percep}}$, and $\lambda_{\text{embed}}$ are balancing parameters. We empirically set $\lambda_{\text{1}}=1$, $\lambda_{\text{2}} = 0.01$, and $\lambda_{\text{3}} = 1000$ in this work.

\section{Quantitative Comparison Supplement}
In previous studies \cite{wang2021dualcam,zhang2022selfdzsr,yue2024kedusr}, it has been commonly observed that training with only the $L_1$ or Charbonnier loss yields superior metrics but worse visual quality. Accordingly, we also re-trained the comparing Dual-camera SR models using only the $L_1$ or Charbonnier loss, denoted as -$\ell$ versions. It is important to note that, since our method include an additional $L_{\mathrm{domain}}$ to supervise the domain-aware embeddings, we use both the Charbonnier loss and domain-aware loss to train {\netname}-$\ell$ model. Tables \ref{tab:dusr_comparison}, \ref{tab:realmcvsr_comparison} and \ref{tab:camera_fusion} lists the quantitative results of the -$\ell$ version models on DuSR-Real, RealMCV-Real, and CameraFusion-Real datasets. It can be observed that our {\netname}-$\ell$ consistently achieves the best performance across most evaluation metrics.

\begin{table}[b]
  \footnotesize
  \centering
  \caption{Quantitative comparisons of -$\ell$ version models the on DuSR-Real Dataset. The best metrics are in bold. }
  \label{tab:dusr_comparison}
  \begin{adjustbox}{width=0.8\linewidth}   
  \begin{tabular}{lccccc}
    \toprule
    \multirow{2}{*}{Method}  & \multicolumn{3}{c}{Full-Image} & Center-Image & Corner-Image \\
    \cmidrule(lr){2-4} \cmidrule(lr){5-5} \cmidrule(lr){6-6}
    & PSNR $\uparrow$ & SSIM $\uparrow$ & LPIPS $\downarrow$ & PSNR/SSIM & PSNR/SSIM \\
    \midrule
 
    DCSR-$\ell$  & 26.25 &  0.8576    &   0.209  &  28.63 / 0.8934    &      25.71 / 0.8454     \\
    SelfDZSR-$\ell$    &  25.71 & 0.8337 & 0.205 &   26.30 / 0.8368  & 25.58 / 0.8326  \\
    KeDuSR-$\ell$   & 27.66 & 0.8890 & 0.177  & \textbf{29.58} / 0.9303 & 27.24  / 0.8750 \\
    \midrule
    \textbf{\netname-$\ell$}  & \textbf{27.79} & \textbf{0.8902} & \textbf{0.173} & 29.50 / \textbf{0.9347} & \textbf{27.41} / \textbf{0.8751} \\
    \bottomrule
  \end{tabular}
  \end{adjustbox}
\end{table}

\begin{table}[h]
  \footnotesize
  \centering
  \caption{Quantitative comparisons of -$\ell$ version models on RealMCVSR-Real. The best metrics are in bold.}
  \label{tab:realmcvsr_comparison}
\begin{adjustbox}{width=0.8\linewidth} 
  \begin{tabular}{lcccccc}
    \toprule
    \multirow{2}{*}{Method} & \multicolumn{3}{c}{Full-Image} & Center-Image & Corner-Image \\
    \cmidrule(lr){2-4} \cmidrule(lr){5-5} \cmidrule(lr){6-6}
    & PSNR $\uparrow$ & SSIM $\uparrow$ & LPIPS $\downarrow$ & PSNR/SSIM & PSNR/SSIM \\
    \midrule
 
    DCSR-$\ell$     &  26.00   &  0.8018   & 0.312 &   27.38 / 0.8315     &  25.67 / 0.7917 \\
    SelfDZSR-$\ell$ & 25.28 & 0.7800  & 0.279  &    25.33 / 0.7746    &      25.33 / 0.7818   \\
    KeDuSR-$\ell$ & 27.05 & 0.8406 & 0.238 & 29.25 / 0.9191 & 26.56 / \textbf{0.8139} \\

    \midrule
    \textbf{\netname-$\ell$} & \textbf{27.11} & \textbf{0.8415} & \textbf{0.236} & \textbf{29.27} / \textbf{0.9254} & \textbf{26.62} / {0.8130} \\
    \bottomrule
  \end{tabular}
  \end{adjustbox}
\end{table}

\begin{table}[h]
  \footnotesize
  \centering
  \caption{Quantitative comparisons of -$\ell$ version models on CameraFusion-Real Dataset. The best metrics are in bold.}
  \label{tab:camera_fusion}
  \begin{adjustbox}{width=0.8\linewidth} 
  \begin{tabular}{lcccccc}
    \toprule
    \multirow{2}{*}{Method} & \multicolumn{3}{c}{Full-Image} & Center-Image & Corner-Image \\
    \cmidrule(lr){2-4} \cmidrule(lr){5-5} \cmidrule(lr){6-6}
    & PSNR $\uparrow$ & SSIM $\uparrow$ & LPIPS $\downarrow$ & PSNR/SSIM & PSNR/SSIM \\
    \midrule
 
    DCSR-$\ell$    &   25.38  &  0.7977  &  0.346  &    26.34  /  0.8106    &     25.17 /   0.7934   \\
    SelfDZSR-$\ell$ & 25.88 & 0.7852 &  0.284 &  26.91 / 0.7960  &  25.66  /  0.7816  \\
    KeDuSR-$\ell$ & 27.53 & 0.8292 & 0.322 & 30.48 / 0.8656 & 26.93 / 0.8169 \\

    \midrule
    \textbf{\netname-$\ell$} & \textbf{27.93} & \textbf{0.8427} & \textbf{0.282} & \textbf{32.10} / \textbf{0.9180} & \textbf{27.16} / \textbf{0.8174} \\
    \bottomrule
  \end{tabular}
  \end{adjustbox}
\end{table}

\begin{figure}[tb]
    \centering
    \includegraphics[width=0.9\linewidth]{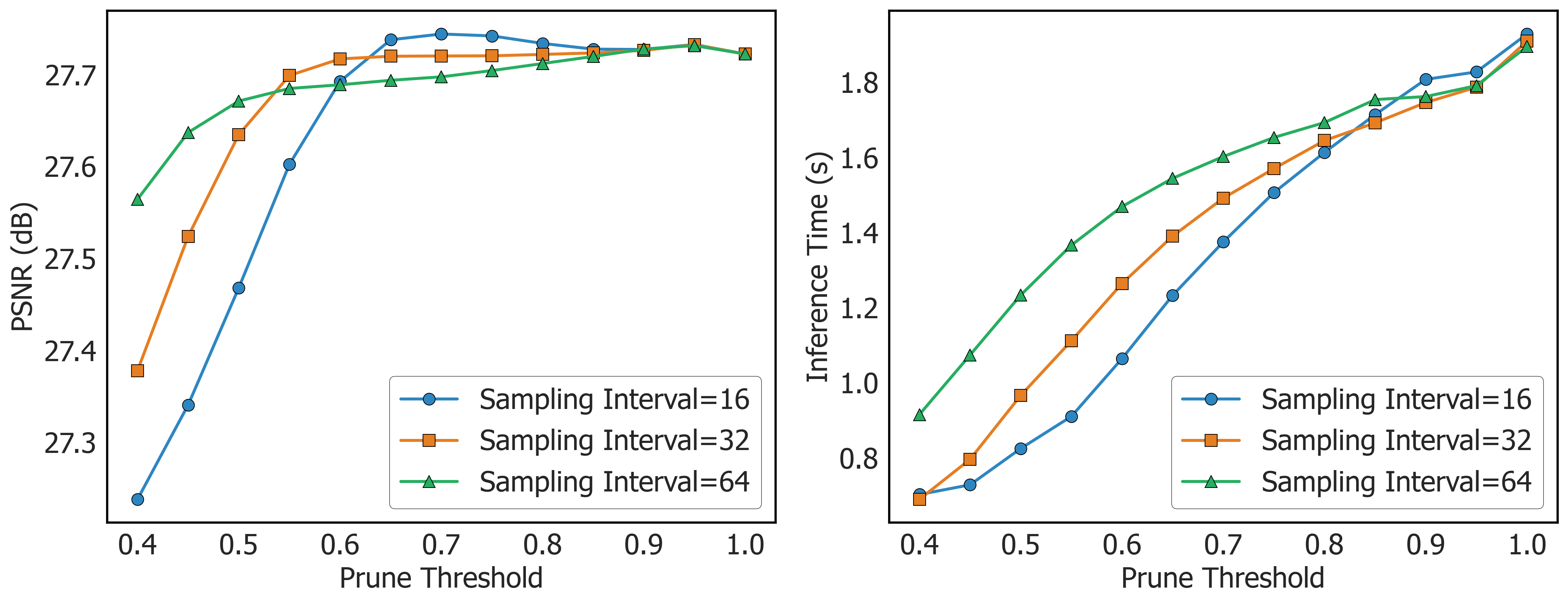}
    \caption{PSNR and inference time in relation to varying thresholds under different sampling intervals. Tests are conducted on DuSR-Real dataset on NVIDA A100 GPU.}
    \label{fig:enter-label}
\end{figure}

\begin{figure}
    \centering
    \includegraphics[width=1\linewidth]{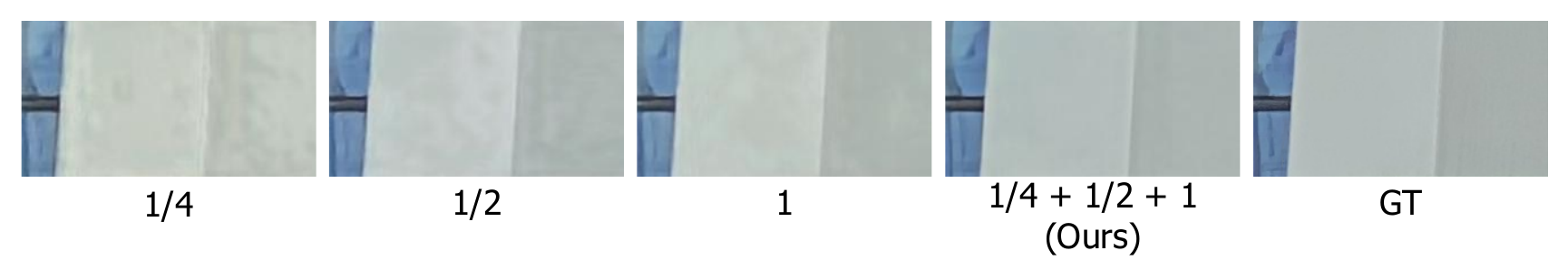}
    \caption{Visual comparisons of ablation study on multi-scale matching. 1/4, 1/2, and 1 denote matching at respective scales.}
    \label{fig:ablationcompare1}
\end{figure}

\begin{figure}
    \centering
    \includegraphics[width=1\linewidth]{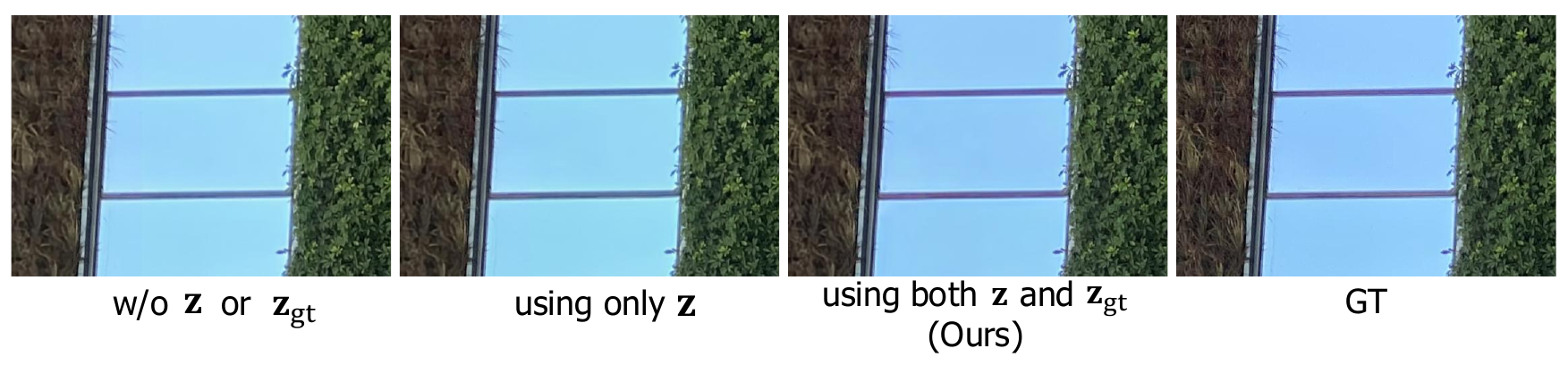}
    \caption{Visual comparisons of ablation study on domain modulation.}
    \label{fig:ablationcompare2}
\end{figure}

\section{More Ablation Study}
\subsection{Ablation on Key Pruning}
We also investigate the impact of the \textbf{sampling interval} and \textbf{threshold} in Key Pruning on both the performance and efficiency of the model. Figure~\ref{fig:enter-label} shows the curves of PSNR and inference time with varying thresholds under different sampling intervals on the DuSR-Real dataset. It can be observed that a higher threshold generally leads to better PSNR but also results in increased inference time. This trend is more evident when the sampling interval is smaller. To balance the model efficiency and performance, we set the sampling interval to 16 and the threshold to 0.7.

\subsection{Visual Comparison}
Fig~\ref{fig:ablationcompare1} presents the visual comparison of the ablation study on multi-scale matching. It is observed that adopting our multi-scaling matching outperforms using  single-scale matching on 1/4, 1/2, or 1 scale. Fig~\ref{fig:ablationcompare2} presents the visual comparison of the ablation study on domain modulation. Our method that uses both $\mathbf{z}$ and $\mathbf{z}_{\rm gt}$ obtains the correct color while the other two models generate color distortion.

\begin{table}[h!]
    \centering
    \caption{Comparison of the model parameters and latency.}
    \begin{adjustbox}{width=0.95\linewidth}
    \begin{tabular}{lcccccccc}
    \toprule
    & SwinIR & Real-ESRGAN  & TTSR & MASA-SR  & DCSR & SelfDZSR & KeDuSR & \netname~ \\
    \midrule
    Params (M)  & 11.75 &  16.70  & 6.25 & 4.02 & 3.19 & 0.52 & 5.63 & 27.42   \\

    Latency (s)  & 4.609 &  0.113  & 7.067 & 6.013 & 1.172 & 0.793 & 0.836 & 1.404   \\
    \bottomrule
    
    \end{tabular}
    \end{adjustbox}
    \label{tab:latency_params}
\end{table}

\section{Complexity Analysis}
We present the latency and number of parameters in Table~\ref{tab:latency_params}. Our method is faster than SwinIR, TTSR, and MASA-SR in terms of latency. Latency indicates the time required to generate one HR result (1792 × 896) using one NVIDIA A100 GPU.

\section{More Visual Comparisons}
\begin{figure}
    \centering
    \includegraphics[width=1\linewidth]{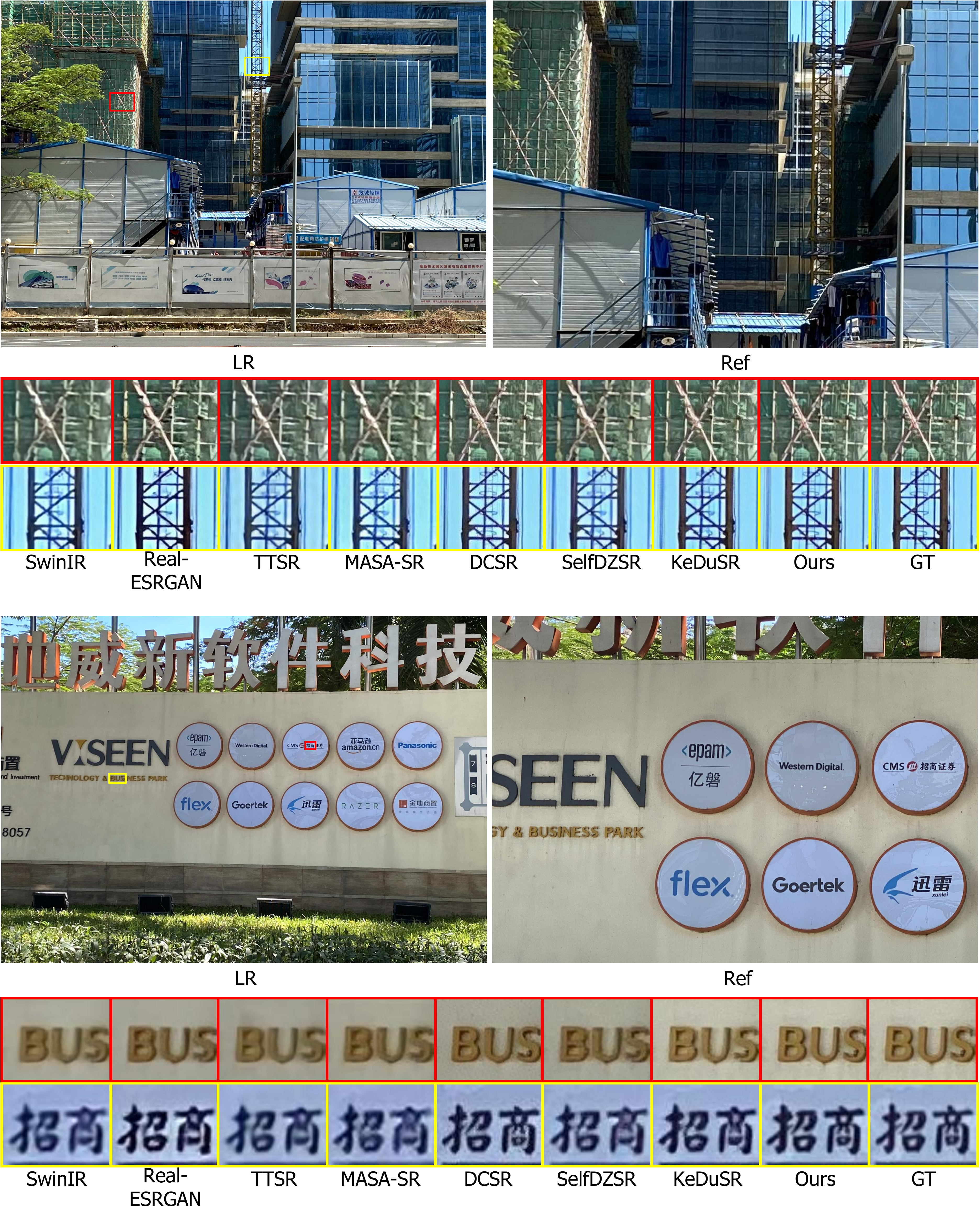}
    \caption{Visual comparisons on CameraFusion-Real.}
    \label{fig:cam_compare_1}
\end{figure}

\begin{figure}
     \centering
     \includegraphics[width=0.9\linewidth]{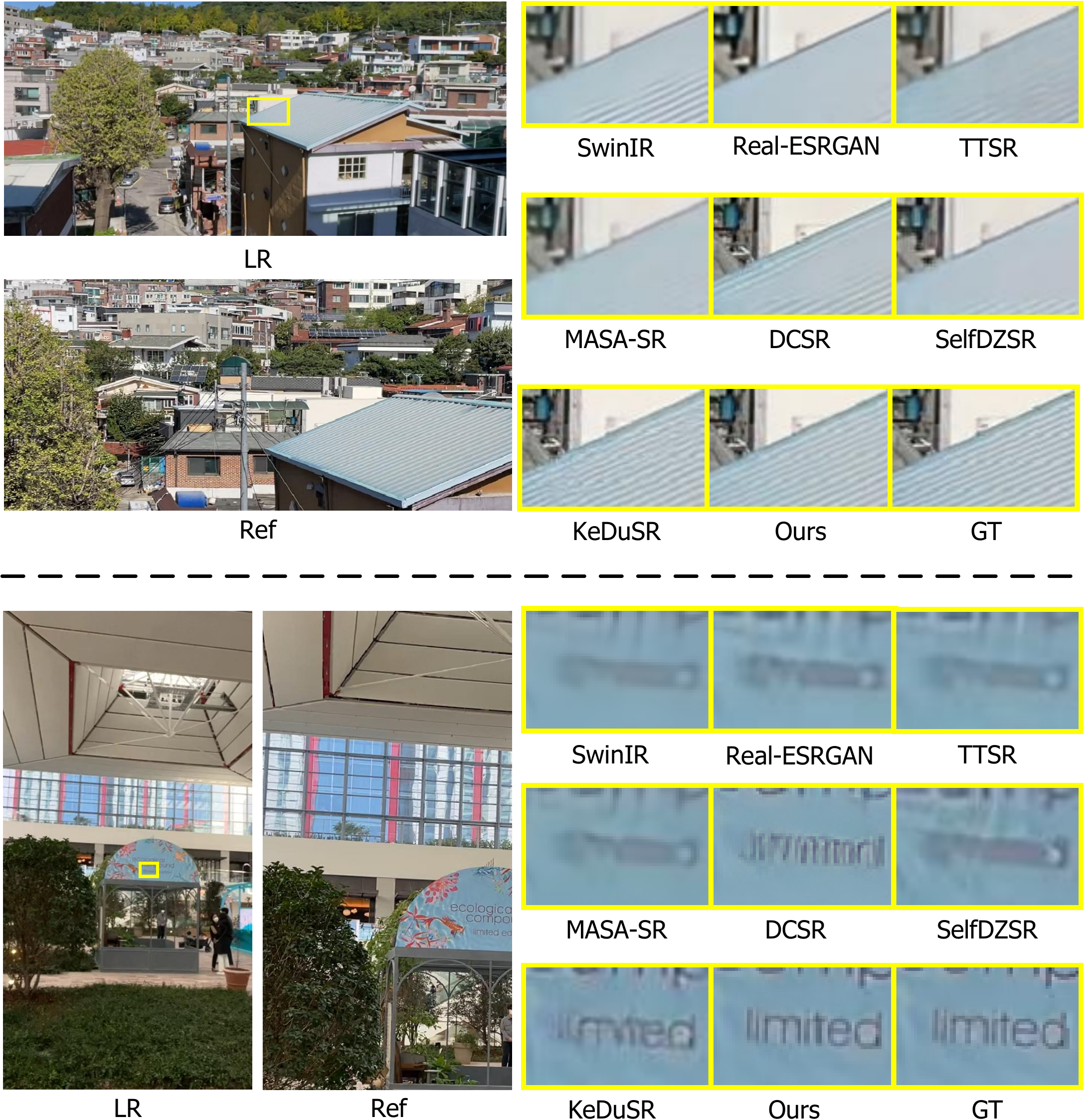}
     \caption{Visual comparisons on RealMCVSR-Real.}
    \label{fig:mcv_compare_1}
\end{figure}

We present more visual comparative results in Figure~\ref{fig:cam_compare_1} and Figure~\ref{fig:mcv_compare_1}. Our method demonstrates superior performance in terms of structure details and color fidelity.

\section{Application and Limitations}
The application of our {\netname} is for computational photography in multi-camera systems such as smartphones, drones, and action cameras. A current limitation of our method lies in the relatively large model parameter size, and its inference speed is not yet real-time. In future work, we will plan to reduce the model parameters and increase the speed.

\clearpage        

\end{document}